%% file: X_ray.tex
\documentclass[journal,twoside,web,runningheads,table]{ieeecolor}
\usepackage{generic}
\usepackage{cite}
\usepackage{amsmath,amssymb,amsfonts}
\usepackage{algorithmic}
\usepackage{graphicx}
\usepackage{textcomp}
\usepackage{xcolor}

\usepackage{pifont}
\newcommand{\cmark}{\ding{51}}%
\newcommand{\xmark}{\ding{55}}%
\usepackage{hhline}
\usepackage{multirow}
\usepackage{soul}
\usepackage{rotating}
\usepackage{cleveref}
\usepackage{float}
\usepackage{dblfloatfix}
\let\labelindent\relax
\usepackage{enumitem}
\usepackage{placeins}
\usepackage{array}
\usepackage{caption}

\def\BibTeX{{\rm B\kern-.05em{\sc i\kern-.025em b}\kern-.08em
    T\kern-.1667em\lower.7ex\hbox{E}\kern-.125emX}}
\begin{document}
\include{file_with_tex_figure_commands}
\title{Recognition of Defective Mineral Wool Using Pruned ResNet Models}

\author{Mehdi Rafiei, Dat Thanh Tran, and Alexandros Iosifidis, \IEEEmembership{Senior Member, IEEE}
\thanks{Mehdi Rafiei and Alexandros Iosifidis are with the Department of Electrical and Computer Engineering, Aarhus University, Denmark (e-mails: rafiei@ece.au.dk, ai@ece.au.dk).}
\thanks{Dat Thanh Tran is with the Department of Computing Sciences, Tampere University, Finland (e-mail: thanh.tran@tuni.fi).}}

\maketitle

\begin{abstract}
Mineral wool production is a non-linear process that makes it hard to control the final quality. Therefore, having a non-destructive method to analyze the product quality and recognize defective products is critical. For this purpose, we developed a visual quality control system for mineral wool. X-ray images of wool specimens were collected to create a training set of defective and non-defective samples. Afterward, we developed several recognition models based on the ResNet architecture to find the most efficient model. In order to have a light-weight and fast inference model for real-life applicability, two structural pruning methods are applied to the classifiers. Considering the low quantity of the dataset, cross-validation and augmentation methods are used during the training. As a result, we obtained a model with more than 98\% accuracy, which in comparison to the current procedure used at the company, it can recognize 20\% more defective products.
\end{abstract}


\section{Introduction}
\label{Sec:Sec1}

\input{Sec/Sec1}

\section{Problem Description}
\label{Sec:Sec2}
\input{Sec/Sec2}

\section{Data collection}
\label{Sec:Sec3}

\input{Sec/Sec3}

\section{Methodology}
\label{Sec:Sec4}
\input{Sec/Sec4}

\section{Model Implementation and Results}
\label{Sec:Sec5}
\input{Sec/Sec5}

\section{Conclusion}
\label{Sec:Sec6}
\input{Sec/Sec6}

\section*{Acknowledgment}

The research leading to the results of this paper received funding from the Innovation Fund Denmark as part of MADE FAST. We thank ROCKWOOL Group company for their help to provide required knowledge regarding the problem and their current methods, and also the support to provide product samples for the test. We thank FORCE Technology company for providing the X-ray lab equipment and their support to do the tests.

\bibliographystyle{ieeetr}
\bibliography{X_ray}

\end{document}

%% file: Sec/Sec1.tex
Mineral wool is one of the most common materials that is used these days as heat insulators and sound absorbers. However, due to the lack of knowledge and limitations in current technology, its manufacturing processes are non-linear, leading to difficulties in achieving full control of the quality of the final product. Therefore, manual quality control strategies at the end of production lines are used in factories. In these strategies, each final product is assessed by an operator to ensure that defective products are rejected. There are two main drawbacks to this process. Firstly, the human-based assessment is inefficient and faces limitations such as slow evaluation, complexity, and subjectivity. The second downside is the inability to detect defects that are not visible on the products' surfaces. Consequently, using X-ray imaging and Computer Vision (CV) tools is considered in this paper as a non-destructive process to assess mineral wool.

The combination of X-ray imaging technology and CV methods has been used in several industrial areas. Casting is one of the areas that uses these methods to visualize the castings' internal structure, assess the products, and detect anomalies \cite{C6}. Several papers frame the casting inspection problem as a binary classification task \cite{C1,C06,C5} to classify X-ray data into defective and non-defective classes. Techniques for binary classification in these works include a weakly supervised Convolutional Neural Network (CNN) model with mutual-channel and cross-entropy loss functions and attention-guided data augmentation in \cite{C1}, a DL-based model made by two sub-networks in \cite{C5}, and a Spatial Attention Bilinear CNN (SA-BCNN) network in \cite{C06}. Also, in the form of a multi-class classification and in aerospace industry applications, several traditional classification methods on casting defect classification of X-ray images of supporting plates in aeromotors are evaluated and compared in \cite{C7}.

Welding is another industrial area that X-ray imaging combined with CV techniques is becoming a common non-destructive assessment method \cite{W13}. In order to have a binary classifier to recognize the defective laser welding, a method composed of a Feed-forward Neural Network (FNN) and a Support Vector Machine (SVM) model was introduced in \cite{W13}. Additionally, several works focus on a multi-class classification of X-ray welding data \cite{W03, W11}. A CNN-based multi-class classification method was proposed in \cite{W03}. In addition, various re-sampling methods were used to tackle the imbalanced class distribution problem. In \cite{W11}, different geometric features (e.g. compactness, symmetry, elongation) were defined to characterize defects in X-ray data. Afterward, the features were considered as inputs to an SVM-based multi-class classifier which divides the problem into $M(M-1)/2$ binary classification problems, where M is the number of classes.

\begin{figure*}[!htbp]
\centering
\includegraphics[width=\textwidth]{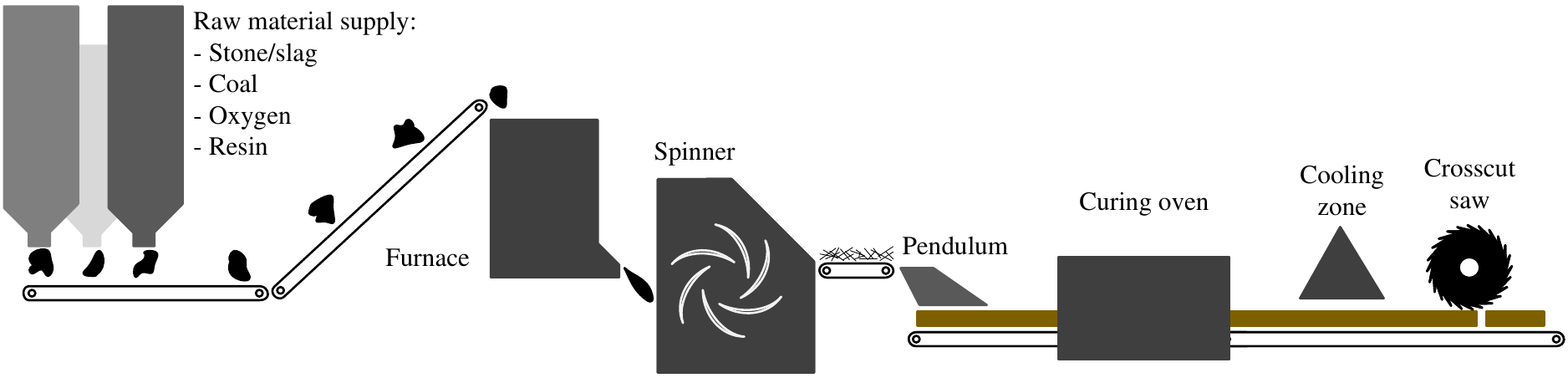}
\caption{Schematic view of the production line.}
\label{Production_line}
\end{figure*}

In this work, we tried to tackle the limitations regarding the defective product recognition of products from an industrial wool production line. In order to have an acceptable solution with a real-life application, we need to develop a model that firstly has a higher accuracy than the current recognition method on the production line, and secondly, it must be lightweight and fast at inference time to make it possible to be run on computationally-restricted hardware. Therefore, we propose to use a neural network architecture with residual connections (ResNet) \cite{ResNet} as the binary classifier. Additionally, two neural network parameter pruning methods (i.e. filter pruning \cite{Fprune} and Asymptotic Soft Filter Pruning (ASFP) \cite{ASFP}) are used to lighten the classifier and accelerate its inference time.

The rest of the paper is structured as follows: In \cref{Sec:Sec2}, the problem is briefly described. The data collection process is illustrated in \cref{Sec:Sec3}. The methods used to implement the binary classifier are shortly presented in \cref{Sec:Sec4}. In \cref{Sec:Sec5}, the implementation process and the related experimental results are provided. Finally, the paper is concluded in \cref{Sec:Sec6}.

%% file: Sec/Sec2.tex
In order to have a better understanding of the problem, a brief description of the production process and the definition of the defect types are presented. Fig.~\ref{Production_line} provides a schematic illustration of a production line of industrial wool. As can be seen, the raw materials, including stones/slags, coal, oxygen, and resin, are melted in a furnace and the melted combination along with binders enter a spinner to make fibers. By use of a pendulum, the produced fibers are placed in layers on a production belt. Afterward, the wool passes through a curing oven and cooling zone, and finally, it is cut by the crosscut saw into the required sizes.



Due to the current limitations in the production line's technical adjustments such as the speed of the spinner, the temperature of the oven, etc., several types of defects can occur in the product. The definitions of the known defects are as follows:

\begin{itemize}
\item\textbf{D1:} drops of melted stones that are not transformed into fibers by the spinner;

\item\textbf{D2:} a bulk of binders that are not distributed on fibers uniformly;

\item\textbf{D3:} a collection of small shots of molten stones;

\item\textbf{D4:} the wool around D1 and D2 defects that is burned due to their heat;

\item\textbf{D5:} a part of wool that is not fully cured in the oven and has moist;

\item\textbf{D6:} dirt from the production line that falls into products.
\end{itemize}

The currently available control system on the production line is an X-ray system that is mainly used for weight per unit area measurement (material distribution), to control the pendulum and the transport speed. However, a thresholding system is adopted on the X-ray system to detect defective products. The threshold values were obtained by trial and error on the production line. By using this thresholding system on several test samples, we concluded that this system cannot recognize all the defects. Therefore, to solve the problem with vision systems and thresholding methods, we propose to use binary-classification CV methods on X-ray data. In addition, to make the developed models suitable for real-life applications, we need to create a light and fast model, while maintaining a good performance.

%% file: Sec/Sec3.tex
\begin{figure}
\centering
\includegraphics[width=\columnwidth]{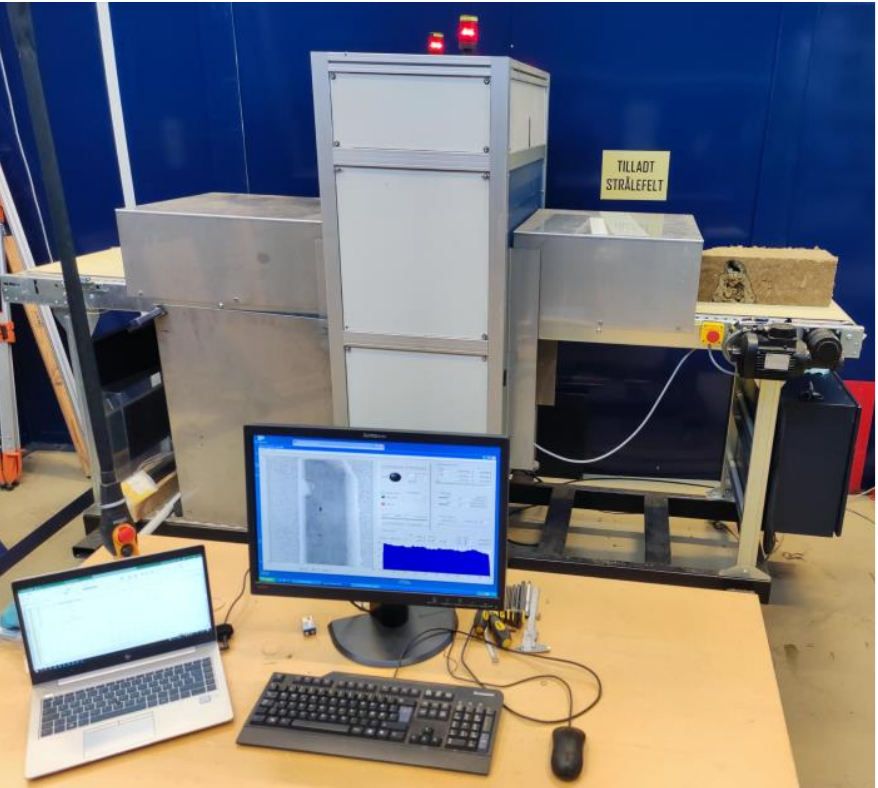}
\caption{X-ray system to collect the data.}
\label{X-ray_systems}
\end{figure}

In order to develop a binary classifier, we need to have a dataset formed by X-ray data of both defective and non-defective products. To this end, we decided to have a test on the same X-ray system that is installed on production lines. Although the mentioned X-ray system is available on several production lines, there is one accessible at the X-ray laboratory of the provider company (Fig.~\ref{X-ray_systems}) that we used to scan product samples. It is a conveyor-based system with adjustable speed, up to 1 m/sec. The X-ray source is placed above the conveyor and a 512 mm line-array detector with 640 pixel elements is placed beneath the top belt. Considering the size of the system, there is a dimension limitation on the samples equal to $1000\times600\times200$ mm ($L\times W\times H$). The setup of the system used for our test was as follows:

\begin{itemize}
    \item Energy:   70 kV;
    \item Flux: 0.5 mA;
    \item Inherent filtration:  approximately 2.5mm Aluminum equivalent.
\end{itemize}

\begin{figure}
\centering
\includegraphics[width=\columnwidth]{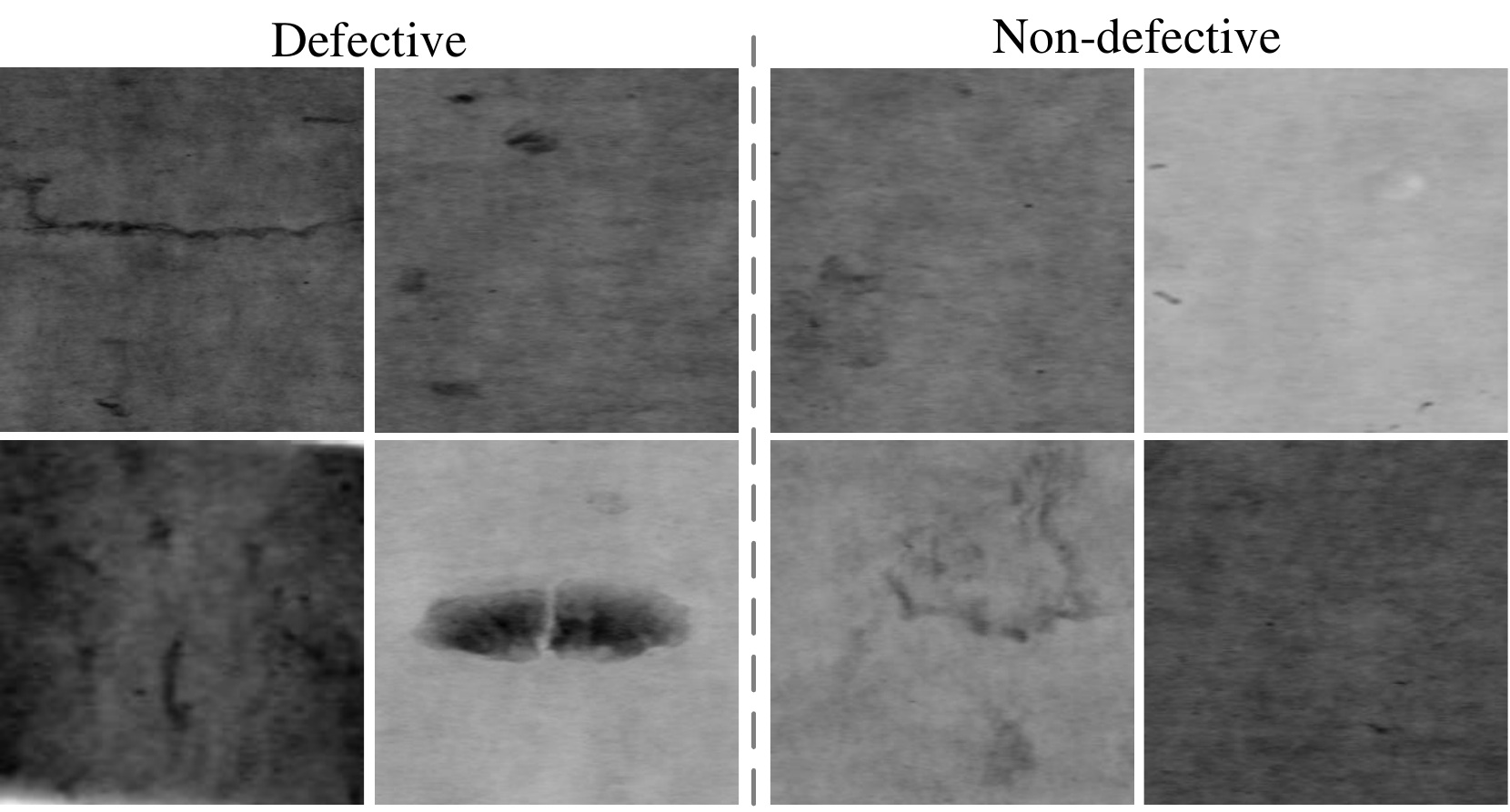}
\caption{Defective and non-defective X-ray samples.}
\label{classifiers_data}
\end{figure}

To perform the test, several defective samples from a few production lines (detected by either using the thresholds, the vision system, or manually by the line operators) were gathered and shipped to the X-ray lab. It is important to mention that each defect can present itself in any orientation, resulting in them appearing differently from various angles on the X-ray data. As the initial data was limited in the number of physical samples, this property was used to increase the volume of data. In other words, when possible, given the size limitations of the X-ray system, each sample was scanned by the X-ray system from 3 different angles. Afterward, the samples were assessed by lab experts to annotate the collected X-ray dataset. Fig.~\ref{classifiers_data} shows a few obtained X-ray samples of defective and non-defective products.

A short report of this test is listed below:
\begin{itemize}
    \item 153 samples were tested that 70 of them were non-defective;
    \item In total, 316 X-ray scans were conducted;
    \item 4 types of defects are visible by X-ray: D1, D2, D3, and D6 (not D4 and D5).
\end{itemize}

As the final dataset for the binary classification, 242 data samples with defects and 236 random samples from the non-defective parts are cropped in total. Since defect types D4 and D5 can not be seen in X-ray imaging, the scanned data related to them are considered non-defective in the ground truth. To prevent having two randomly cropped data from the same source image (with potential overlapping) in training and validation datasets, each source image is only used to have one cropped data sample. The images are in PNG format and with a final resolution of 244×244 pixels.

%% file: Sec/Sec4.tex
\subsection{Classifier}

By considering the impact of using deeper neural networks and their ability to learn better the information in a dataset, one would wonder whether it would be easy to train and employ deeper neural networks for the task at hand. Clearly, we need to consider the obstacles to increasing the depth of a neural network, such as vanishing/exploding gradients, which are solved by available techniques like normalized initialization and intermediate normalization layers. Even after solving the obstacles, the authors of \cite{ResNet} stated that a degradation problem has occurred in deeper networks that saturates the accuracy. It should be considered that this degradation is not caused by overfitting.

The reason behind the mentioned problem is the difficulty of optimization of highly deep networks \cite{ResNet}. To tackle the problem, the deep residual learning framework is introduced that adds an identifying mapping known as "skip connection" to the layers. This new architecture results in a higher ability to train deeper networks in comparison to networks without the skip connection. ResNet shows high potential in image recognition tasks \cite{wen2022new, dong2021deep}. Several architectures of ResNet are available with diverse depths, such as ResNet-18, ResNet-34, etc.

\subsection{Neural network parameter pruning}

Although a wider and deeper neural network might lead to learning better the information in the data and achieving higher accuracy, such an over-parameterized network leads to high computational cost and requires high memory. In this regard, neural network parameter pruning is a well-suited solution to lighten the size of deep neural networks and improve their inference time. We adopted two filter pruning techniques.

\begin{figure*}
\centering
\includegraphics[width=0.85\textwidth]{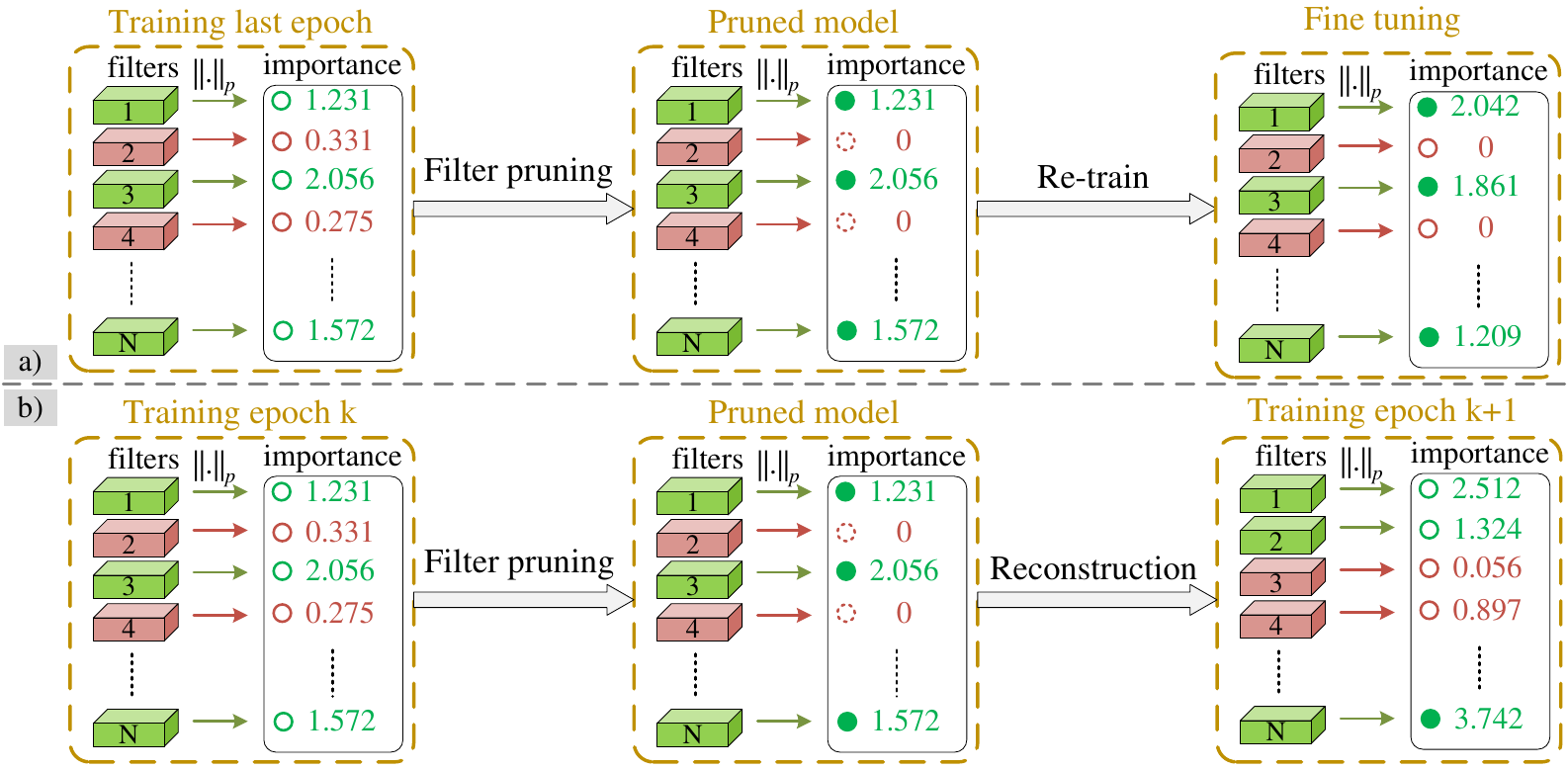}
\caption{Overview of a) filter pruning, and b) ASFP processes.}
\label{prune}
\end{figure*}

The first one \cite{Fprune}, is a typical filter pruning method, which contains training, pruning, and fine-tuning (retraining) steps. In the pruning step, a large number of filters are fixed to zero and they are not retrained during the fine-tuning step. Fig.~\ref{prune}-a illustrates the last training epoch, pruning, and fine-tuning steps. As it can be seen, after the last epoch of training, the $l_p$-norm is calculated for each filter as the filter selection criterion to prune the least effective filters. Afterward, in the filter pruning stage, the method sets the filters with lower $l_p$-norms equal to zero. The number of selected filters to be pruned is obtained based on the pruning rate provided by the user. Then, the model would be fine-tuned without updating the pruned filters. 

Not retraining the pruned filters during the fine-tuning step leads to unrecoverable information loss and consequently reduction of the optimization space. To solve this drawback, the Asymptotic Soft Filter Pruning (ASFP) method is introduced in \cite{ASFP}. The main concepts of this method are the updating of the pruned filters during training and an asymptotic way of pruning the neural network. As a consequence, there is no reduction in the optimization space, and the information of the training set is gradually concentrated in the remaining filters, respectively. Fig.~\ref{prune}-b illustrates an epoch of training in the ASFP method. As can be seen, the process after each training epoch is like the pruning after training in the previous method and then, the model would be trained by the pruned filters to be updated at the next epoch. It must be mentioned that the number of selected filters to be pruned is determined based on the pruning rate of each epoch, and it starts from a small number at the beginning and increases to reach to the required pruning rate at the last epoch.

\subsection{Cross-validation}

Cross-validation is a common technique to evaluate an ML model and assess its performance. It eases the comparison and selection of a suitable model for a predictive modeling problem. Although there are several strategies to apply cross-validation on a model (such as Hold-out, \textit{k}-folds, Leave-one-out, etc.), all of them share the same algorithm as below:

\begin{enumerate}
    \item divide the dataset into train and validation subsets (based on the selected cross-validation technique),
    \item train and validate the model by those subsets,
    \item repeat the previous steps \textit{N} times (\textit{N} depends on the selected technique).
\end{enumerate}

In order to implement the cross-validation on our data in the classification task, we used \textit{k}-fold strategy. \textit{k}-fold cross-validation helps to see the performance of the suggested model on the whole dataset, while the validation data is not seen in the training phase. After choosing the value \textit{k} (can be any number less than the number of samples in the dataset), the following steps must be taken (Fig.~\ref{Cross-val}):

\begin{figure}
\centering
\includegraphics[width=\columnwidth]{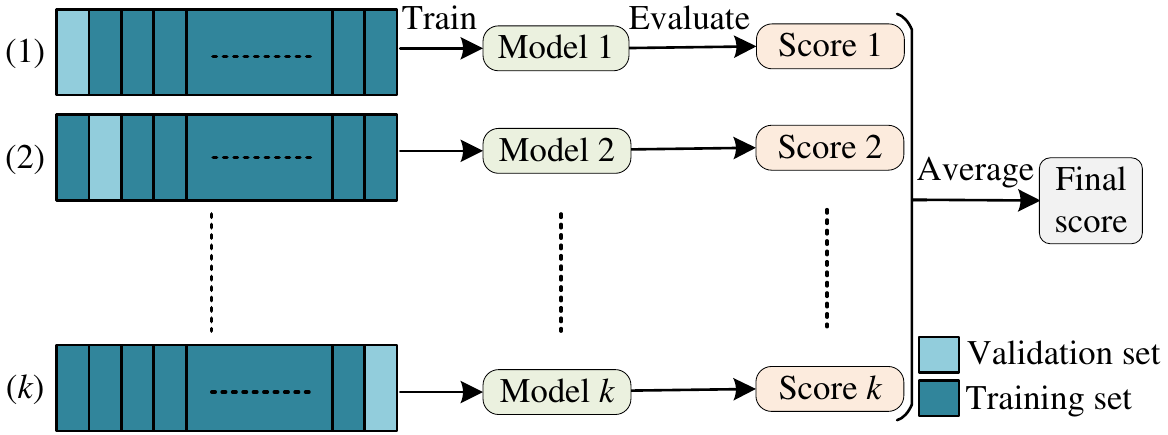}
\caption{\textit{k}-fold cross-validation training scheme.}
\label{Cross-val}
\end{figure}

%% file: Sec/Sec5.tex
\subsection{Implementation} 

In this section, a brief description of the classifier implementation and related considerations are provided.

For the binary classification, ResNet architectures (i.e. ResNet-18, 34, 50, 101, and 152) are trained and their performances are analyzed. In order to assess the impact of using pre-trained models, for all the architectures, two scenarios including the pre-trained models on ImageNet dataset \cite{imagenet} and training the model from scratch, are assessed. In order to reduce the impact of low data quality and also have an effective method to assess the models' performances, the \textit{k}-fold cross-validation technique is used and the \textit{k} value is set equal to 10. Additionally, random cropping and random flipping data augmentations are used to enrich the dataset. Also, Adam and Stochastic Gradient Descent (SGD) are considered as optimizers to assess their impact on the model's performance. Besides these, a reducing learning rate is used to improve the convergence of the training processes.

In order to assess parameter pruning efficiency on the trained models, both filter pruning and ASFP methods are considered for all the scenarios. For filter pruning method, 3 rounds of pruning are applied for each model with pruning rates of $p_1=p_2=p_3=30\%$ (as it is suggested in the original paper \cite{Fprune}). ASFP method is used to train each model 3 times with 3 different pruning rates (i.e. $p=10$, $20$, or $30\%$).

Furthermore, in order to have more reliable results, each training process is applied 10 times and the mean and standard deviation of the metrics are calculated and reported.

\subsection{Results}

\subsubsection{Classification results}

The results related to the classification are presented in Table~\ref{binary_results}. It can be seen that using the pre-trained weights leads to better performances in comparison to training the model from scratch on our dataset. The performance obtained by using SGD or Adam optimizers is similar and in some cases (ResNet-50 and 101) Adam shows a slightly better performance, while in other cases SGD is better. For all model depths, the highest validation accuracy is between 98.25 and 98.96\%. ResNet-101 achieves the best performance, but we cannot observe a linear relation between the depth of the model and its accuracy.


\input{Tables/binary_results}

\input{Tables/prune_results}


As it was mentioned before, there is the possibility to detect the defects by using a thresholding-based process. Those thresholds vary based on the mean area weight of the products. However, to reduce the size of the collected samples and transportation costs, most of them were cut into small pieces around the defects. Therefore, for most of the test samples, it was not possible to calculate the average area weight of the whole product and consequently use the correct threshold. However, by applying all the thresholds on the test data, we could reach a recognition accuracy of the defective products between 76\% and 82\%. Therefore, it can be seen that the binary classifier has at least 16.96\% higher accuracy (20\% more detection of defective products).

\subsubsection{Pruning results}

The results related to the binary classification with both neural network parameter pruning methods described in \cref{Sec:Sec4} are presented in Table~\ref{prune_results}. The table shows the average accuracy of baseline models and the pruned ones, along with the corresponding accuracy drops. It can be seen that pruning impacted the validation accuracy of the models. While using the filter pruning method decreased the accuracy by a maximum of 1.34\% in some cases (except for some cases such as ResNet-34 with SGD optimizer and using pre-trained weights), it slightly increased the accuracy in other cases. On the other hand, ASFP leads to higher accuracy drops. By average, it can be seen that, the regular filter pruning method provides more acceptable results in the studied problem. Moreover, when considering higher pruning rates, the results obtained by using filter pruning are more stable. This shows that filter pruning has the ability to better retain performance compared to ASFP in the studied problem. 


In addition, to illustrate the impact of filter pruning on the models' memory usage and required inference computation time, a comparison between the baseline model and pruned ones is presented in Table~\ref{parameters}. In this table, the number of parameters and the Floating Point Operators (FLOPs) are listed for each baseline model. The effect of applying each of the three pruning rounds on the parameter count and FLOPs is also provided. The number of FLOPs indicates the number of operations needed for classifying an input image using a model. Therefore, a lower number of parameters indicate a lighter model in term of memory usage, and, a lower number of FLOPs indicates a faster model during inference. The results show that, in total, the models are lighten between 40 and 10\%, and their number of operations are also reduced to between 45 and 10\% of the baselines' inference time. However, for cases of ResNet-18 and 34, pruning reduces the number of parameters and FLOPs much more than for other architectures. Considering the results shown in Tables~\ref{prune_results} and~\ref{parameters}, ResNet-18 with 3 rounds of filter pruning is the most suitable model for the problem, as it leads to a good accuracy versus model size trade-off.

\input{Tables/Parameters}

%% file: Tables/binary_results.tex

\begin{table}
\captionsetup{justification=centering}
\caption{Comparison of defective product recognition results gained by different model depths, optimizers, and initial weighs.}
\label{binary_results}
\centering
\scriptsize
\begin{tabular}{>{\centering\arraybackslash}p{0.9cm}>{\centering\arraybackslash}p{1.2cm}>{\centering\arraybackslash}p{1.3cm}>{\centering\arraybackslash}p{1.0cm}>{\centering\arraybackslash}p{2.3cm}}
\hline\hline
\rowcolor[HTML]{A6A6A6} 
\cellcolor[HTML]{A6A6A6}                               & \cellcolor[HTML]{A6A6A6}                            & \cellcolor[HTML]{A6A6A6}                              & \multicolumn{2}{l}{\cellcolor[HTML]{A6A6A6}Baseline   validation accuracy (\%)} \\ \hhline{*{3}{~}*{2}{-}}  
\rowcolor[HTML]{A6A6A6} 
\multirow{-2}{*}{\cellcolor[HTML]{A6A6A6}\begin{tabular}[c]{@{}l@{}}ResNet \\Depth\end{tabular}} & \multirow{-2}{*}{\cellcolor[HTML]{A6A6A6}\begin{tabular}[c]{@{}l@{}}Optimizer\end{tabular}} & \multirow{-2}{*}{\cellcolor[HTML]{A6A6A6}\begin{tabular}[c]{@{}l@{}}Pre-trained\end{tabular}} & Mean                               & Standard deviation                               \\ \hline\hline
\rowcolor[HTML]{FFFFFF} 
\cellcolor[HTML]{FFFFFF}                               & \cellcolor[HTML]{FFFFFF}                            & \xmark                                 & 98.08                                 & 0.24                                    \\ \hhline{*{2}{~}*{3}{-}}  
\cellcolor[HTML]{FFFFFF}                               & \multirow{-2}{*}{\cellcolor[HTML]{FFFFFF}Adam}      & \cellcolor[HTML]{FFFFFF}\cmark         & 98.33                                 & 0.27                                    \\ \hhline{*{1}{~}*{4}{-}}  
\rowcolor[HTML]{FFFFFF} 
\cellcolor[HTML]{FFFFFF}                               & \cellcolor[HTML]{FFFFFF}                            & \xmark                                 & 97.28                                 & 0.23                                    \\ \hhline{*{2}{~}*{3}{-}}  
\rowcolor[HTML]{FFFFFF} 
\multirow{-4}{*}{\cellcolor[HTML]{FFFFFF}18}           & \multirow{-2}{*}{\cellcolor[HTML]{FFFFFF}SGD}       & \cmark                                 & 98.92                                 & 0.15                                    \\ \hline
\rowcolor[HTML]{F2F2F2} 
\cellcolor[HTML]{F2F2F2}                               & \cellcolor[HTML]{F2F2F2}                            & \xmark                                 & 98.02                                 & 0.12                                    \\ \hhline{*{2}{~}*{3}{-}}  
\rowcolor[HTML]{F2F2F2} 
\cellcolor[HTML]{F2F2F2}                               & \multirow{-2}{*}{\cellcolor[HTML]{F2F2F2}Adam}      & \cmark                                 & 98.28                                 & 0.20                                    \\ \hhline{*{1}{~}*{4}{-}}  
\rowcolor[HTML]{F2F2F2} 
\cellcolor[HTML]{F2F2F2}                               & \cellcolor[HTML]{F2F2F2}                            & \xmark                                 & 97.44                                 & 0.36                                    \\ \hhline{*{2}{~}*{3}{-}}  
\rowcolor[HTML]{F2F2F2} 
\multirow{-4}{*}{\cellcolor[HTML]{F2F2F2}34}           & \multirow{-2}{*}{\cellcolor[HTML]{F2F2F2}SGD}       & \cmark                                 & 98.44                                 & 0.21                                    \\ \hline
\rowcolor[HTML]{FFFFFF} 
\cellcolor[HTML]{FFFFFF}                               & \cellcolor[HTML]{FFFFFF}                            & \xmark                                 & 98.29                                 & 0.27                                    \\ \hhline{*{2}{~}*{3}{-}}  
\rowcolor[HTML]{FFFFFF} 
\cellcolor[HTML]{FFFFFF}                               & \multirow{-2}{*}{\cellcolor[HTML]{FFFFFF}Adam}      & \cmark                                 & 98.37                                 & 0.31                                    \\ \hhline{*{1}{~}*{4}{-}}  
\rowcolor[HTML]{FFFFFF} 
\cellcolor[HTML]{FFFFFF}                               & \cellcolor[HTML]{FFFFFF}                            & \xmark                                 & 96.44                                 & 0.26                                    \\ \hhline{*{2}{~}*{3}{-}}  
\rowcolor[HTML]{FFFFFF} 
\multirow{-4}{*}{\cellcolor[HTML]{FFFFFF}50}           & \multirow{-2}{*}{\cellcolor[HTML]{FFFFFF}SGD}       & \cmark                                 & 98.25                                 & 0.41                                    \\ \hline
\rowcolor[HTML]{F2F2F2} 
\cellcolor[HTML]{F2F2F2}                               & \cellcolor[HTML]{F2F2F2}                            & \xmark                                 & 98.12                                 & 0.00                                    \\ \hhline{*{2}{~}*{3}{-}}  
\rowcolor[HTML]{F2F2F2} 
\cellcolor[HTML]{F2F2F2}                               & \multirow{-2}{*}{\cellcolor[HTML]{F2F2F2}Adam}      & \cmark                                 & 98.96                                 & 0.30                                    \\ \hhline{*{1}{~}*{4}{-}}  
\rowcolor[HTML]{F2F2F2} 
\cellcolor[HTML]{F2F2F2}                               & \cellcolor[HTML]{F2F2F2}                            & \xmark                                 & 91.12                                 & 2.50                                    \\ \hhline{*{2}{~}*{3}{-}}  
\rowcolor[HTML]{F2F2F2} 
\multirow{-4}{*}{\cellcolor[HTML]{F2F2F2}101}          & \multirow{-2}{*}{\cellcolor[HTML]{F2F2F2}SGD}       & \cmark                                 & 98.75                                 & 0.30                                    \\ \hline
                                                       &                                                     & \xmark                                 & 98.02                                 & 0.27                                    \\ \hhline{*{2}{~}*{3}{-}}  
                                                       & \multirow{-2}{*}{Adam}                              & \cmark                                 & 98.44                                 & 0.27                                    \\ \hhline{*{1}{~}*{4}{-}}  
                                                       &                                                     & \xmark                                 & 82.89                                 & 2.87                                    \\ \hhline{*{2}{~}*{3}{-}}  
\multirow{-4}{*}{152}                                  & \multirow{-2}{*}{SGD}                               & \cmark                                 & 98.65                                 & 0.21                                    \\ \hline\hline
\end{tabular}
\end{table}

%% file: Tables/prune_results.tex
\begin{table*}
\captionsetup{justification=centering}
\caption{\label{prune_results} Overall performance of pruning ResNet for different model depth.}
\centering
\footnotesize
\resizebox{\textwidth}{!}{
\begin{tabular}{>{\centering\arraybackslash}p{0.45cm}>{\centering\arraybackslash}p{1.1cm}>{\centering\arraybackslash}p{0.5cm}>{\centering\arraybackslash}p{1.4cm}>{\centering\arraybackslash}p{0.7cm}>{\centering\arraybackslash}p{0.7cm}>{\centering\arraybackslash}p{0.7cm}>{\centering\arraybackslash}p{0.7cm}>{\centering\arraybackslash}p{0.7cm}>{\centering\arraybackslash}p{0.8cm}>{\centering\arraybackslash}p{0.7cm}>{\centering\arraybackslash}p{0.8cm}>{\centering\arraybackslash}p{0.7cm}>{\centering\arraybackslash}p{0.8cm}>{\centering\arraybackslash}p{0.7cm}>{\centering\arraybackslash}p{0.8cm}}
\hline\hline
\rowcolor[HTML]{A6A6A6} 
\multicolumn{3}{c}{\cellcolor[HTML]{A6A6A6}}                                                                                                                 & \cellcolor[HTML]{A6A6A6}                                                      & \multicolumn{6}{c}{\cellcolor[HTML]{A6A6A6}Filter pruning}                                                                                                      & \multicolumn{6}{c}{\cellcolor[HTML]{A6A6A6} ASFP}                                                                                                                                        \\ \hhline{*{4}{~}*{12}{-}}  
\rowcolor[HTML]{A6A6A6} 
\multicolumn{3}{c}{\cellcolor[HTML]{A6A6A6}}                                                                                                                 & \cellcolor[HTML]{A6A6A6}                                                      & \multicolumn{2}{c}{\cellcolor[HTML]{A6A6A6}Round 1} & \multicolumn{2}{c}{\cellcolor[HTML]{A6A6A6}Round 2} & \multicolumn{2}{c}{\cellcolor[HTML]{A6A6A6}Round 3} & \multicolumn{2}{c}{\cellcolor[HTML]{A6A6A6}\begin{tabular}[c]{@{}c@{}}Prune rate\\ = 0.1\end{tabular}} & \multicolumn{2}{c}{\cellcolor[HTML]{A6A6A6}\begin{tabular}[c]{@{}c@{}}Prune rate\\ = 0.2\end{tabular}} & \multicolumn{2}{c}{\cellcolor[HTML]{A6A6A6}\begin{tabular}[c]{@{}c@{}}Prune rate\\ = 0.3\end{tabular}} \\ \hhline{*{4}{~}*{12}{-}}  
\rowcolor[HTML]{A6A6A6} 
\multicolumn{3}{c}{\multirow{-3}{*}{\cellcolor[HTML]{A6A6A6}\begin{tabular}[c]{@{}p{2.25cm}@{}}ResNet Depth\\ (Optimizer)\\ “Pre-trained”\end{tabular}}} & \multirow{-3}{*}{\cellcolor[HTML]{A6A6A6}\begin{tabular}[c]{@{}c@{}}Baseline\\ validation \\ accuracy \\(\%)\end{tabular}} & Val. Acc.                & Acc. drops               & Val. Acc.                & Acc. drops               & Val. Acc.                & Acc. drops               & Val. Acc.                    & Acc. drops                   & Val. Acc.                    & Acc. drops                   & Val. Acc.                    & Acc. drops                   \\ \hline\hline
\rowcolor[HTML]{FFFFFF} 
\cellcolor[HTML]{FFFFFF}                                        & \cellcolor[HTML]{FFFFFF}                                           & “\xmark”                   & 98.08                                                                         & 98.16                    & 0.08                     & 98.12                    & 0.04                     & 98.5                     & 0.42                     & 96.91                        & -1.17                        & 96.45                        & -1.63                        & 94.05                        & -4.03                        \\ \hhline{*{2}{~}*{14}{-}}  
\rowcolor[HTML]{FFFFFF} 
\cellcolor[HTML]{FFFFFF}                                        & \multirow{-2}{*}{\cellcolor[HTML]{FFFFFF}(Adam)}                   & “\cmark”                   & 98.33                                                                         & 98.58                    & 0.25                     & 98.75                    & 0.42                     & 98.74                    & 0.41                     & 99.08                        & 0.75                         & 99.34                        & 1.01                         & 98.51                        & 0.18                         \\ \hhline{*{1}{~}*{15}{-}}  
\rowcolor[HTML]{FFFFFF} 
\cellcolor[HTML]{FFFFFF}                                        & \cellcolor[HTML]{FFFFFF}                                           & “\xmark”                   & 97.28                                                                         & 97.74                    & 0.46                     & 97.49                    & 0.21                     & 97.46                    & 0.18                     & 88.8                         & -8.48                        & 86.14                        & -11.1                       & 84.72                        & -12.6                       \\ \hhline{*{2}{~}*{14}{-}}  
\rowcolor[HTML]{FFFFFF} 
\multirow{-4}{*}{\cellcolor[HTML]{FFFFFF}18}                    & \multirow{-2}{*}{\cellcolor[HTML]{FFFFFF}(SGD)}                    & “\cmark”                   & 98.92                                                                         & 97.62                    & -1.3                     & 97.58                    & -1.34                    & 97.58                    & -1.34                    & 98.8                         & -0.12                        & 99.37                        & 0.45                         & 98.02                        & -0.9                         \\ \hline
\rowcolor[HTML]{F2F2F2} 
\cellcolor[HTML]{F2F2F2}                                        & \cellcolor[HTML]{F2F2F2}                                           & “\xmark”                   & 98.02                                                                         & 98.05                    & 0.03                     & 98.33                    & 0.31                     & 98.4                     & 0.38                     & 97.77                        & -0.25                        & 96.85                        & -1.17                        & 97.25                        & -0.77                        \\ \hhline{*{2}{~}*{14}{-}}  
\rowcolor[HTML]{F2F2F2} 
\cellcolor[HTML]{F2F2F2}                                        & \multirow{-2}{*}{\cellcolor[HTML]{F2F2F2}(Adam)}                   & “\cmark”                   & 98.28                                                                         & 98.47                    & 0.19                     & 98.68                    & 0.4                      & 98.61                    & 0.33                     & 99.22                        & 0.94                         & 98.84                        & 0.56                         & 98.85                        & 0.57                         \\ \hhline{*{1}{~}*{15}{-}}  
\rowcolor[HTML]{F2F2F2} 
\cellcolor[HTML]{F2F2F2}                                        & \cellcolor[HTML]{F2F2F2}                                           & “\xmark”                   & 97.44                                                                         & 97.64                    & 0.2                      & 97.7                     & 0.26                     & 97.7                     & 0.26                     & 87.67                        & -9.77                        & 88.12                        & -9.32                        & 88.68                        & -8.76                        \\ \hhline{*{2}{~}*{14}{-}}  
\rowcolor[HTML]{F2F2F2} 
\multirow{-4}{*}{\cellcolor[HTML]{F2F2F2}34}                    & \multirow{-2}{*}{\cellcolor[HTML]{F2F2F2}(SGD)}                    & “\cmark”                   & 98.44                                                                         & 97.63                    & -0.81                    & 94.85                    & -3.59                    & 79.35                    & -19.1                   & 98.87                        & 0.43                         & 99.3                         & 0.86                         & 98.62                        & 0.18                         \\ \hline
\rowcolor[HTML]{FFFFFF} 
\cellcolor[HTML]{FFFFFF}                                        & \cellcolor[HTML]{FFFFFF}                                           & “\xmark”                   & 98.29                                                                         & 98.54                    & 0.25                     & 98.86                    & 0.57                     & 98.62                    & 0.33                     & 94.57                        & -3.72                        & 94.23                        & -4.06                        & 92.94                        & -5.35                        \\ \hhline{*{2}{~}*{14}{-}}  
\rowcolor[HTML]{FFFFFF} 
\cellcolor[HTML]{FFFFFF}                                        & \multirow{-2}{*}{\cellcolor[HTML]{FFFFFF}(Adam)}                   & “\cmark”                   & 98.37                                                                         & 98.47                    & 0.1                      & 98.37                    & 0                        & 98.46                    & 0.09                     & 99.1                         & 0.73                         & 98.93                        & 0.56                         & 97.69                        & -0.68                        \\ \hhline{*{1}{~}*{15}{-}}  
\rowcolor[HTML]{FFFFFF} 
\cellcolor[HTML]{FFFFFF}                                        & \cellcolor[HTML]{FFFFFF}                                           & “\xmark”                   & 96.44                                                                         & 97.07                    & 0.63                     & 97.35                    & 0.91                     & 97.28                    & 0.84                     & 83.67                        & -12.8                       & 82.08                        & -14.4                       & 79.07                        & -17.4                       \\ \hhline{*{2}{~}*{14}{-}}  
\rowcolor[HTML]{FFFFFF} 
\multirow{-4}{*}{\cellcolor[HTML]{FFFFFF}50}                    & \multirow{-2}{*}{\cellcolor[HTML]{FFFFFF}(SGD)}                    & “\cmark”                   & 98.25                                                                         & 98.54                    & 0.29                     & 98.36                    & 0.11                     & 98.54                    & 0.29                     & 98.98                        & 0.73                         & 99.08                        & 0.83                         & 97.74                        & -0.51                        \\ \hline
\rowcolor[HTML]{F2F2F2} 
\cellcolor[HTML]{F2F2F2}                                        & \cellcolor[HTML]{F2F2F2}                                           & “\xmark”                   & 98.12                                                                         & 98.54                    & 0.42                     & 98.61                    & 0.49                     & 98.61                    & 0.49                     & 93.58                        & -4.54                        & 94.99                        & -3.13                        & 92.87                        & -5.25                        \\ \hhline{*{2}{~}*{14}{-}}  
\rowcolor[HTML]{F2F2F2} 
\cellcolor[HTML]{F2F2F2}                                        & \multirow{-2}{*}{\cellcolor[HTML]{F2F2F2}(Adam)}                   & “\cmark”                   & 98.96                                                                         & 98.54                    & -0.42                    & 98.75                    & -0.21                    & 98.61                    & -0.35                    & 99.26                        & 0.3                          & 99.05                        & 0.09                         & 99.42                        & 0.46                         \\ \hhline{*{1}{~}*{15}{-}}  
\rowcolor[HTML]{F2F2F2} 
\cellcolor[HTML]{F2F2F2}                                        & \cellcolor[HTML]{F2F2F2}                                           & “\xmark”                   & 91.12                                                                         & 96.51                    & 5.39                     & 96.93                    & 5.81                     & 97.35                    & 6.23                     & 87.96                        & -3.16                        & 85.28                        & -5.84                        & 84.4                         & -6.72                        \\ \hhline{*{2}{~}*{14}{-}}  
\rowcolor[HTML]{F2F2F2} 
\multirow{-4}{*}{\cellcolor[HTML]{F2F2F2}101}                   & \multirow{-2}{*}{\cellcolor[HTML]{F2F2F2}(SGD)}                    & “\cmark”                   & 98.75                                                                         & 98.26                    & -0.49                    & 98.47                    & -0.28                    & 98.54                    & -0.21                    & 98.16                        & -0.59                        & 99.12                        & 0.37                         & 98.94                        & 0.19                         \\ \hline
                                                                &                                                                    & “\xmark”                   & 98.02                                                                         & 98.8                     & 0.78                     & 98.8                     & 0.78                     & 98.89                    & 0.87                     & 92.94                        & -5.08                        & 91.95                        & -6.07                        & 93.51                        & -4.51                        \\ \hhline{*{2}{~}*{14}{-}}  
                                                                & \multirow{-2}{*}{(Adam)}                                           & “\cmark”                   & 98.44                                                                         & 98.54                    & 0.1                      & 98.59                    & 0.15                     & 98.61                    & 0.17                     & 99.26                        & 0.82                         & 97.38                        & -1.06                        & 97.62                        & -0.82                        \\ \hhline{*{1}{~}*{15}{-}}  
                                                                &                                                                    & “\xmark”                   & 82.89                                                                         & 87.31                    & 4.42                     & 92.06                    & 9.17                     & 95.19                    & 12.3                     & 84.32                        & 1.43                         & 83.64                        & 0.75                         & 81.98                        & -0.91                        \\ \hhline{*{2}{~}*{14}{-}}  
\multirow{-4}{*}{152}                                           & \multirow{-2}{*}{(SGD)}                                            & “\cmark”                   & 98.65                                                                         & 98.33                    & -0.32                    & 98.54                    & -0.11                    & 98.4                     & -0.25                    & 98.97                        & 0.32                         & 99.13                        & 0.48                         & 98.52                        & -0.13                        \\ \hline\hline
\end{tabular}}
\end{table*}

%% file: Tables/Parameters.tex
\begin{table*}
\captionsetup{justification=centering}
\caption{\label{parameters} Filter pruning impact on models' memory usage and inference computation time.}
\centering
\footnotesize
\begin{tabular}{>{\centering\arraybackslash}p{1.0cm}>{\centering\arraybackslash}p{1.6cm}>{\centering\arraybackslash}p{1.1cm}>{\centering\arraybackslash}p{1.6cm}>{\centering\arraybackslash}p{1.1cm}>{\centering\arraybackslash}p{1.6cm}>{\centering\arraybackslash}p{1.1cm}>{\centering\arraybackslash}p{1.6cm}>{\centering\arraybackslash}p{1.1cm}}
\hline\hline
\rowcolor[HTML]{A6A6A6} 
\cellcolor[HTML]{A6A6A6}                                 & \multicolumn{2}{c}{\cellcolor[HTML]{A6A6A6}}                                 & \multicolumn{6}{c}{\cellcolor[HTML]{A6A6A6}Filter pruning}                                                                                                      \\ \hhline{*{3}{~}*{6}{-}}
\rowcolor[HTML]{A6A6A6} 
\cellcolor[HTML]{A6A6A6}                                 & \multicolumn{2}{c}{\multirow{-2}{*}{\cellcolor[HTML]{A6A6A6}\begin{tabular}[c]{@{}p{2.6cm}@{}}Baseline model\end{tabular}}} & \multicolumn{2}{c}{\cellcolor[HTML]{A6A6A6}Round 1} & \multicolumn{2}{c}{\cellcolor[HTML]{A6A6A6}Round 2} & \multicolumn{2}{c}{\cellcolor[HTML]{A6A6A6}Round 3} \\ \hhline{*{1}{~}*{8}{-}}
\rowcolor[HTML]{A6A6A6} 
\multirow{-3}{*}{\cellcolor[HTML]{A6A6A6}\begin{tabular}[c]{@{}c@{}}ResNet \\Depth\end{tabular}} & Parameters                            & FLOPs                              & parameters               & FLOPs                  & parameters               & FLOPs                  & parameters               & FLOPs                  \\\hline\hline
\rowcolor[HTML]{FFFFFF} 
18                                                       & 1.2e+7                                & 1.8e+9                           & 4.8e+6                   & 1.0e+9               & 2.0e+6                   & 6.3e+8               & 9.0e+5                   & 4.1e+8               \\\hline
\rowcolor[HTML]{F2F2F2} 
34                                                       & 2.2e+7                                & 3.7e+9                           & 5.3e+6                   & 1.3e+9               & 1.4e+6                   & 6.2e+8               & 5.0e+5                   & 3.5e+8               \\\hline
\rowcolor[HTML]{FFFFFF} 
50                                                       & 2.6e+7                                & 4.1e+9                           & 1.8e+7                   & 3.1e+9               & 1.4e+7                   & 2.4e+9               & 1.1e+7                   & 2.0e+9               \\\hline
\rowcolor[HTML]{F2F2F2} 
101                                                      & 4.5e+7                                & 7.8e+9                           & 3.2e+7                   & 5.8e+9               & 2.3e+7                   & 4.4e+9               & 1.8e+7                   & 3.4e+9               \\\hline
152                                                      & 6.0e+7                                & 1.2e+10                           & 4.3e+7                   & 8.5e+9               & 3.2e+7                   & 6.3e+9               & 2.4e+7                   & 4.9e+9             \\ \hline\hline
\end{tabular}
\end{table*}

%% file: Sec/Sec6.tex
In this paper, to improve the accuracy of recognizing defective mineral wool products achieved by the utilized thresholding method, we propose to use computer vision techniques on X-ray data. We collected a dataset by carrying out an X-ray test on wool specimens and then we used ResNet models along with two structured neural network parameter pruning methods to achieve a highly accurate model while lowering its parameter count and the number of operations. Considering the maximum 82\% accuracy obtained by following the existing process based on thresholding, the proposed model can improve the recognition process by reaching an accuracy of more than 98\%. These results indicate that we can recognize 20\% more defective products compared to the thresholding method. Moreover, the use of filter pruning led to models' memory usage reduction between 40 and 10\% and a reduction in the number of operations from 2.5 to 10 times, while retaining their accuracy levels. These results show the high potential of the model to be used as a real-life solution condition to recognize and reject defective products.
